\documentclass[sigconf]{acmart}
\usepackage{booktabs}
\usepackage{amsmath}
\usepackage{algorithm}
\usepackage{algpseudocode}
\usepackage[font=footnotesize,caption=false]{subfig}

\setcopyright{rightsretained}

\acmConference[HIP'17]{The 4th International Workshop on Historical Document Imaging and Processing}{November 2017}{Kyoto, Japan} 
\acmYear{2017}
\copyrightyear{2017}

\acmPrice{00.00}

\begin{document}
\title{\systemname: Page Boundary Extraction in Historical Handwritten Documents}

\author{Chris Tensmeyer}
\orcid{0000-0002-0761-3243}
\affiliation{%
  \institution{Brigham Young University}
  \city{Provo} 
  \state{Utah}
  \country{USA}
  \postcode{84604}
}
\email{tensmeyer@byu.edu}

\author{Brian Davis}
\affiliation{%
  \institution{Brigham Young University}
  \city{Provo} 
  \state{Utah}
  \country{USA}
  \postcode{84604}
}
\email{briandavis@byu.net}

\author{Curtis Wigington}
\affiliation{%
  \institution{Brigham Young University}
  \city{Provo} 
  \state{Utah} 
  \country{USA}
  \postcode{84604}
}
\email{wigington@byu.net}

\author{Iain Lee}
\affiliation{%
  \institution{Brigham Young University}
  \city{Provo} 
  \state{Utah}
  \country{USA}
  \postcode{84604}
}
\email{iclee141@byu.net}

\author{Bill Barrett}
\affiliation{%
  \institution{Brigham Young University}
  \city{Provo} 
  \state{Utah}
  \country{USA}
  \postcode{84604}
}
\email{barrett@cs.byu.edu}

\renewcommand{\shortauthors}{C. Tensmeyer et al.}
\newcommand{\systemname}{PageNet}
\newcommand{\squeezeup}{\vspace{-2.5mm}}
\newcommand{\BiState}[2]{%
  \State{\makebox[1.5cm]{#1\hfill}#2}%
  }

\begin{abstract}
When digitizing a document into an image, it is common to include a surrounding border region to visually indicate that the entire document is present in the image.
However, this border should be removed prior to automated processing.
In this work, we present a deep learning based system, \systemname, which identifies the main page region in an image in order to segment content from both textual and non-textual border noise.
In \systemname{}, a Fully Convolutional Network obtains a pixel-wise segmentation which is post-processed into the output quadrilateral region.
We evaluate \systemname{} on 4 collections of historical handwritten documents and obtain over 94\% mean intersection over union on all datasets and approach human performance on 2 of these collections.
Additionally, we show that \systemname{} can segment documents that are overlayed on top of other documents.
\end{abstract}



\maketitle

\section{Introduction}
\label{sec:intro}

\begin{figure}
\subfloat[\systemname{} Segmentation]{\includegraphics[height=0.25\textwidth]{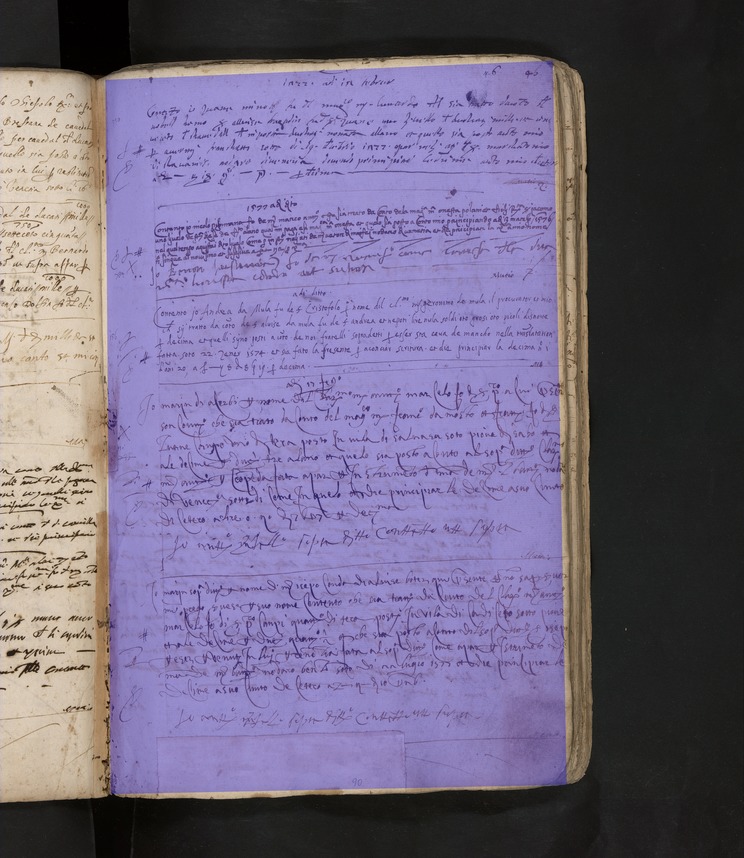}}
\hspace{5pt}
\subfloat[GrabCut Segmentation ]{\includegraphics[height=0.25\textwidth]{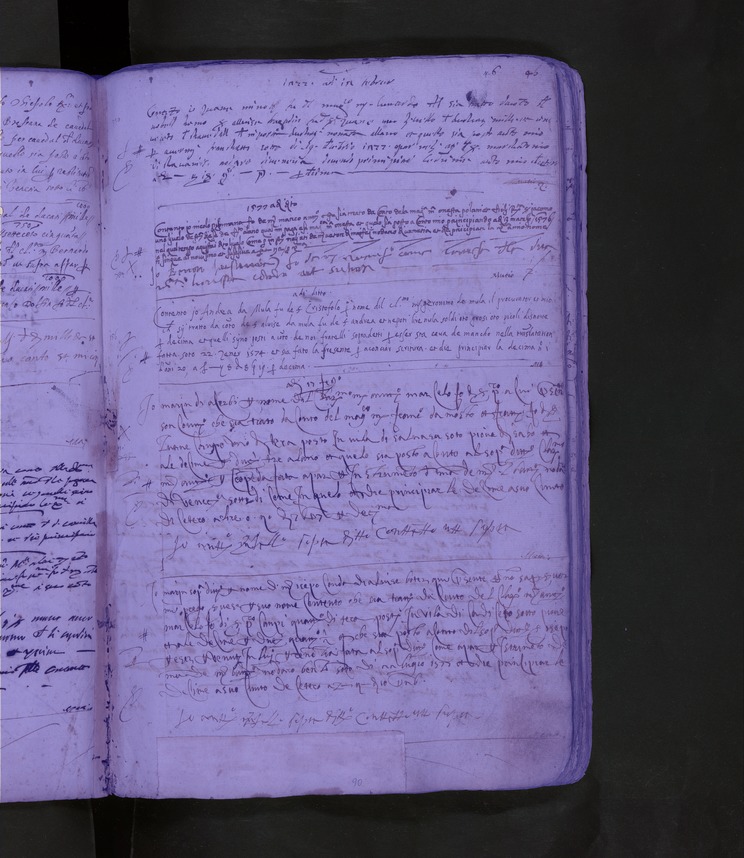}}

\caption{\systemname{}, our proposed system, segments the main page region from border noise such as book edges and portions of the opposite page.  Traditional segmentation algorithms like GrabCut struggle with some types of border noise, though are effective at removing the background.}
\label{fig:example_seg}
\end{figure}

\begin{figure*}
\subfloat[]{\includegraphics[width=0.19\textwidth,height=0.24\textwidth]{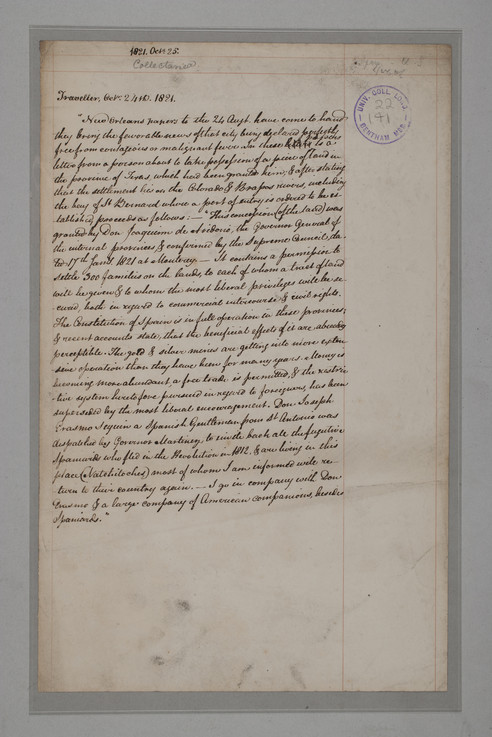}} \hspace{3pt}
\subfloat[]{\includegraphics[width=0.19\textwidth,height=0.24\textwidth]{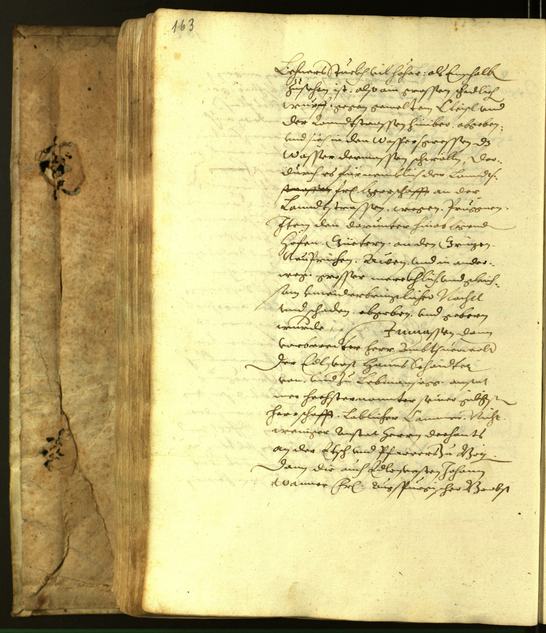}}
\hspace{3pt}
\subfloat[]{\includegraphics[width=0.19\textwidth,height=0.24\textwidth]{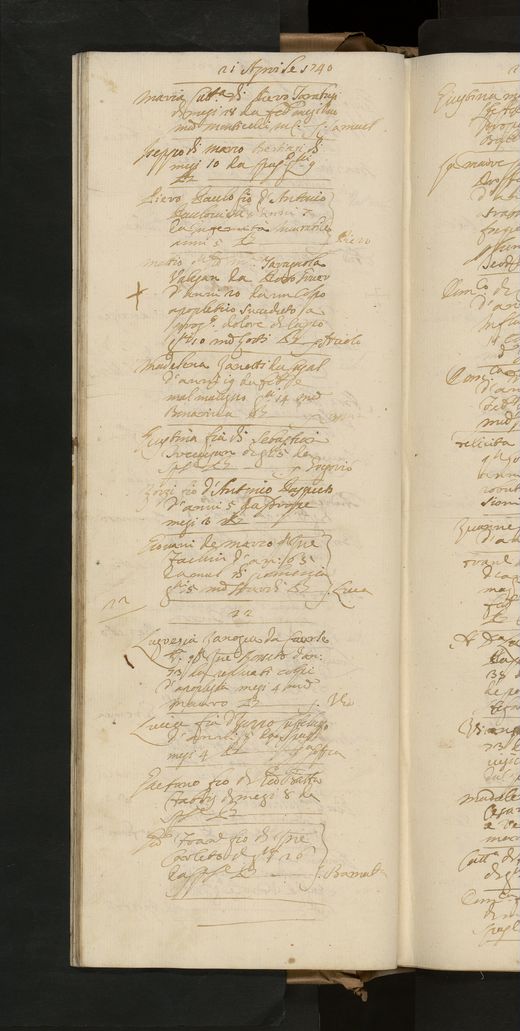}} \hspace{3pt}
\subfloat[]{\includegraphics[width=0.19\textwidth,height=0.24\textwidth]{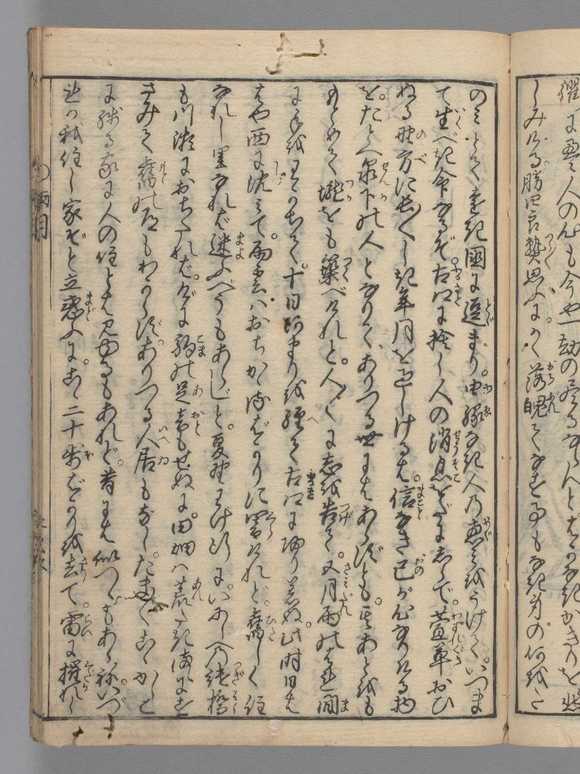}}
\hspace{3pt}
\subfloat[]{\includegraphics[width=0.19\textwidth,height=0.24\textwidth]{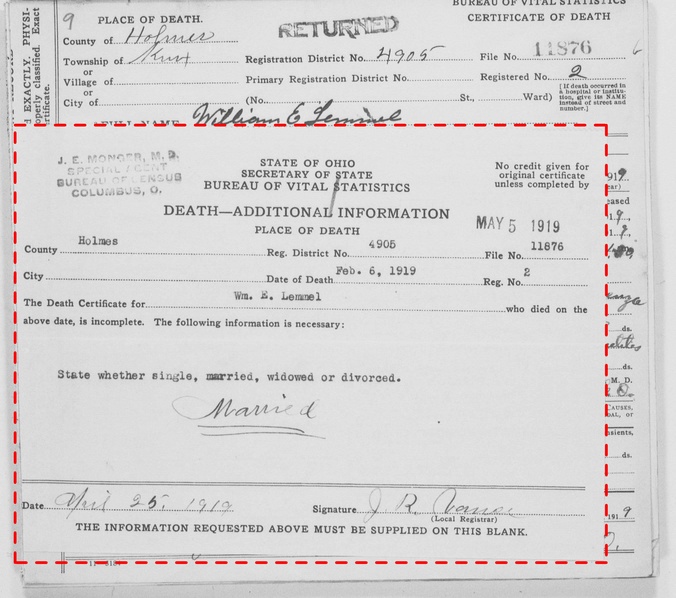}}

\caption{Examples of common border noise.  (a-d) Background around the border. (b-d) Book edges.  (c-d) Textual noise due to neighboring pages. (e) Textual noise due to one document (red square) overlayed on top of another.}
\label{fig:example_problems}
\end{figure*}

When digitizing a document into an image, it is common to include a surrounding border region to visually indicate that the entire document is present in the image.
The \emph{border noise} resulting from this process can interfere with document analysis algorithms and cause undesirable results.
For example, text may be detected and transcribed outside of the main page region when textual noise is present due to a partially visible neighboring page.
Thus, it is beneficial to preprocess the image to remove the border region.

Some examples of border noise in historical documents include background, book edges, overlayed documents, and parts of neighboring pages (see Figure~\ref{fig:example_problems}).
While removing background is feasible using simple segmentation techniques, other types of border noise are more challenging.

A number of approaches have been developed to remove border noise (e.g.,~\cite{fan2002,shafait2009,stamatopoulos2010,bukhari2011,chakraborty2016b}).
However, as noted in~\cite{chakraborty2016a}, many prior methods make assumptions that do not necessarily hold for historical handwritten documents.
These assumptions about the input image include consistent text size, absolute location of border noise, straight text lines, and distances between page text and border~\cite{chakraborty2016a}.

In this work, we propose \emph{\systemname}, a deep learning based system that, given an input image, predicts a bounding quadrilateral for the \emph{main page region} (see Figure~\ref{fig:example_seg}).
\systemname{} is composed of a Fully Convolutional Network (FCN) and post-processing.
The FCN component outputs a pixel-wise segmentation which is post-processed to extract a quadrilateral shaped region.
Detecting the main page region is similar to finding the \emph{page frame}~\cite{shafait2009}, but the former task detects the entire page while the latter crops the detected region to the page text.
\systemname{} is able to robustly handle a variety of border noise because, as a learning based system, it does not make any explicit assumptions about the border noise, layout, or content of the input image.
Though learning methods can potentially overfit the training collection, we demonstrate that \systemname{} generalizes to other collections of documents.

We experimentally evaluated \systemname{} on 5 collections of historical documents that we manually annotated with quadrilateral regions.
To estimate human agreement on this task, we made a second set of annotations for a subset of images.
On our primary test-set, \systemname{} achieves 97.4\% mean Intersection over Union (mIoU) while the human agreement is 98.3\% mIoU.
On all collections tested, we achieve a mIoU of $>$94\%.
Additionally, we show that \systemname{} is capable of segmenting documents that are overlayed on top of other documents.
In order to support the reproducibility of our work, we are releasing our code, models, and dataset annotations, available at \url{www.example.com}.

\section{Related Work}
\label{sec:related_work}

We take a segmentation approach to removing border noise, so we review the literature on traditional border noise removal techniques and on segmentation.

Fan et al. remove non-textual marginal scanning noise (e.g., large black regions) from printed documents by detecting noise with a resolution reduction approach and removing noise through region growing or local thresholding~\cite{fan2002}.
Shafait et al. handle both textual and non-textual marginal noise by finding the page frame of an image that maximizes a quality function w.r.t. an input layout analysis composed of sets of connected components, text lines, and zones~\cite{shafait2008}.
The method of Shafait and Bruel examines the local densities of black and white pixels in fixed image regions to identify noise and also removes connected components near the image border~\cite{shafait2009}.
Stamatopoulos et al. proposed a system based on projection profiles to find the two individual page frames in images of books where two pages are shown.  
They report an average F-measure of 99.2\% on 15 historical books~\cite{stamatopoulos2010}.
Bukhari et al. find the page frame for camera captured documents by detecting text lines, aligning the text line end points, and estimating a straight line from the endpoints using RANdom SAmple Consensus (RANSAC) linear regression~\cite{bukhari2011}.
For further reading, we refer the reader to a recent survey on border noise removal~\cite{chakraborty2016a}.

Another formulation of the border noise removal problem is to find the four corners of the bounding quadrilateral of the page, which is a sub-task shared with perspective dewarping techniques.
Jagannathan et al. find page corners in camera captured documents by identifying two sets of parallel lines and two sets of perpendicular lines in the perspective transformed image~\cite{jagannathan2005}.
Yang et al. use a Hough line transform to detect boundaries in binarized images of license plates~\cite{yang2012}.

\begin{figure*}
\subfloat[Image]{\includegraphics[height=0.22\textwidth,width=0.22\textwidth]{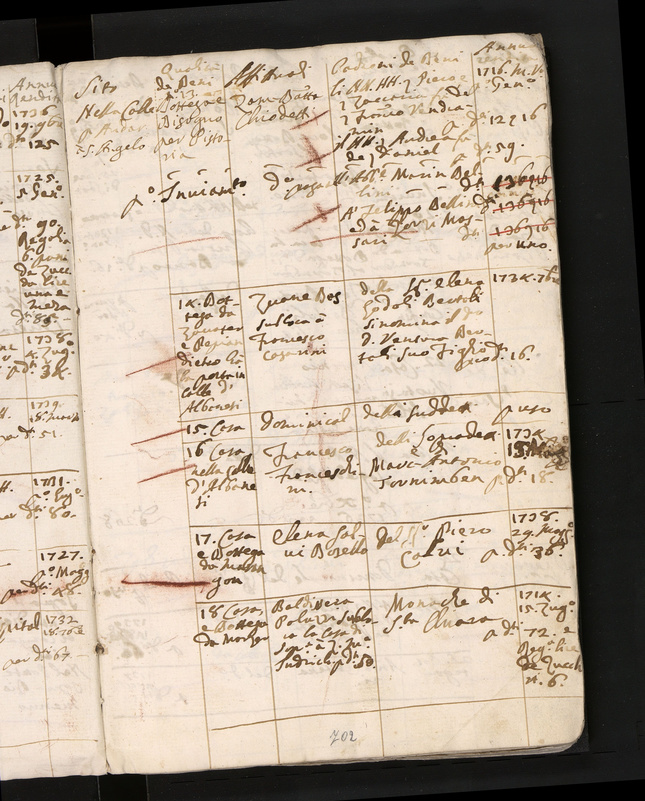}} \hspace{5pt}
\subfloat[FCN Prediction]{\includegraphics[height=0.22\textwidth] {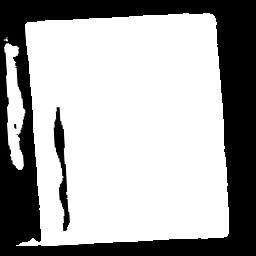}} \hspace{5pt}
\subfloat[Largest Connected Component]{\includegraphics[height=0.22\textwidth]{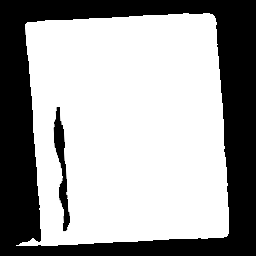}} \hspace{5pt}
\subfloat[Fill Holes]{\includegraphics[height=0.22\textwidth]{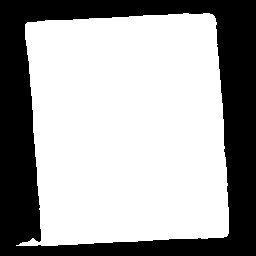}} 

\subfloat[Oriented Bounding Rectangle]{\includegraphics[height=0.22\textwidth]{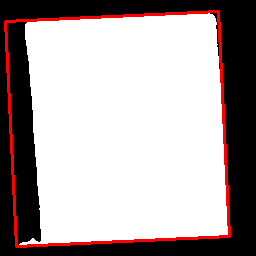}} \hspace{5pt}
\subfloat[Quadrilateral Minimizing IoU]{\includegraphics[height=0.22\textwidth]{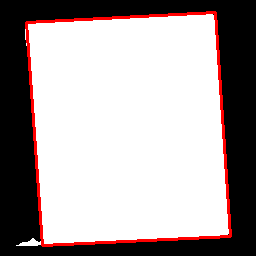}} \hspace{5pt}
\subfloat[Result]{\includegraphics[height=0.22\textwidth,width=0.22\textwidth]{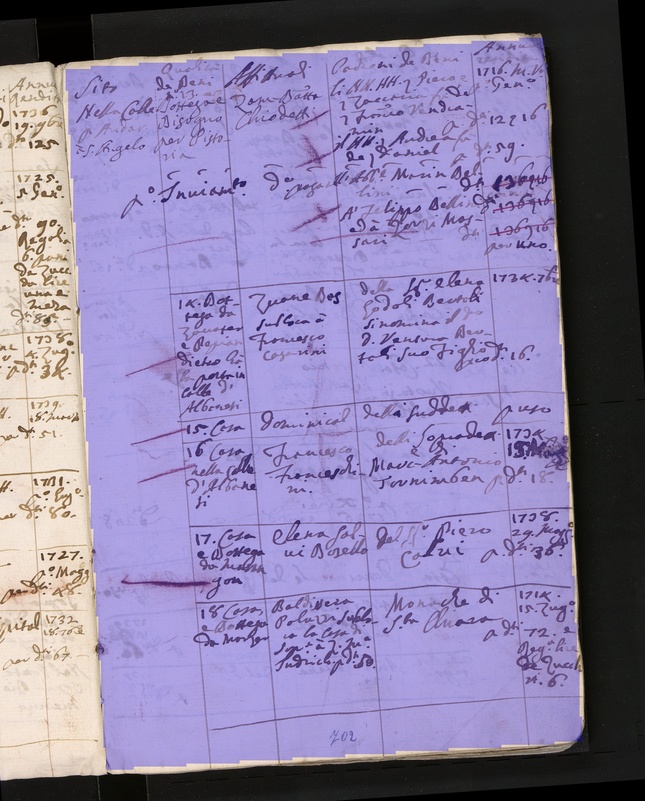}}

\caption{Post processing to extract quadrilateral from FCN predictions.}
\label{fig:post_processing}
\end{figure*}

Intelligent scissors segments an object from background by finding a least cost path through a weighted graph defined over pixels, subject to input constraints~\cite{mortensen1995}.
Active Contours or Snakes formulate segmentation as a continuous optimization problem and finds object boundaries by minimizing a boundary energy cost and the cost of deformation from some prior shape~\cite{kass1987}.
Graph Cut methods (e.g.\cite{boykov2001}) formulate image segmentation as finding the minimum cut over a graph constructed from the image.
Weights in the graph are determined by pixel colors and by per-pixel apriori costs of being assigned to the foreground and background segments.
GrabCut iteratively performs graph cut segmentations, starting from an initial rough bounding box.
The result of each iteration is used to refine a color model which is used to construct the edge weights for the next iteration~\cite{rother2004}.

Several neural network approaches have been proposed for image segmentation.
The Fully Convolution Network (FCN) learns an end-to-end classification function for each pixel in the image~\cite{long2015}.
However, the output is poorly localized due to downsampling in the FCN architecture used.
Zheng et al. integrated a Conditional Random Field (CRF) graphical model into the FCN to improve segmentation localization~\cite{zheng2015}.
In contrast, our approach maintains the input resolution and therefore does not suffer from poor localization.
The Spatial Transformer Network~\cite{jaderberg2015} learns a latent affine transformation in conjunction with a task specific objective, effectively learning cropping, rotation, and skew correct in an end-to-end fashion.
In our case, we are interested in directly learning a pre-processing transformation from ground truth.
Chen and Seuret used a convolutional network to classify super pixels as background, text, decoration, and comments~\cite{chen2017}. 
Super pixel based features would not work in our case as neighboring pages are identical to the main page region in local appearance.

\section{Method}
\label{sec:method}

In this section, we describe~\systemname{}, which takes in a document image and outputs the coordinates of four corners of the quadrilateral that encloses the main page region.
\systemname{} has two parts: 
\begin{enumerate}
\item A Fully Convolutional Neural Network (FCN) to classify pixels as page or background
\item Post processing to extract a quadrilateral region
\end{enumerate}

\subsection{Pixel Classification}
\label{sec:fcn}

FCNs are learning models that alternate convolutions with element-wise non-linear functions~\cite{long2015}.
They differ from traditional CNNs (e.g., AlexNet) that have fully connected layers which constrain the input image size.
In particular, the FCN used in ~\systemname{} maps an input RGB image $x \in \mathbb{R}^{3 \times H \times W} \rightarrow y \in \mathbb{R}^{H \times W}$, where $y_{ij} \in [0,1]$ is the probability that pixel $x_{ij}$ is part of the main page region.

Each layer of a basic FCN performs the operation
\begin{equation} \label{eq:fcn}
x_{\ell} = g(W_{\ell} \ast x_{\ell - 1} + b_{\ell})
\end{equation}
where $\ell$ is a layer index, $\ast$ indicates multi-channel 2D convolution, $W_{\ell}$ is a set of learnable convolution filters, $b_{\ell}$ is a learnable bias term, and $g$ is a non-linear function.
In our case, we use $g(z) = \operatorname{ReLU}(z) = \max(0, z)$ as element-wise non-linear rectification.
In some FCN architectures, spatial resolution is decreased at certain layers through pooling or strided convolution and increased through bilinear interpolation or backwards convolution~\cite{long2015}.
In the last layer of the network, $\operatorname{ReLU}$ is replaced with a channel-wise softmax operation (over 2 channels in our case) to obtain output probabilities for each pixel.

\systemname{} uses a successful multi-scale FCN architecture originally designed for binarizing handwritten text~\cite{tensmeyer2017}.
This FCN operates on 4 image scales: $1, \frac{1}{2}, \frac{1}{4}, \frac{1}{8}$.
The full resolution image scale uses 7 sequential layers (Eq.~\ref{eq:fcn}), and each smaller layer uses one layer less than the previous (e.g. $\frac{1}{8}$ scale uses 4 layers).
The input feature maps of the 3 smallest scales are obtained by $2\times 2$ average-pooling over the output of the first layer of the next highest image scale.
The FCN concatenates the output feature maps of each scale, upsampling them to the input image size using bilinear interpolation.
Thus the architecture both preserves the original input resolution and benefits from a larger context window obtained through downsampling.
This is followed by 2 more convolution layers and the softmax operation.
For full details on the FCN architecture, we refer the reader to~\cite{tensmeyer2017}.
We performed initial experiments (not shown) with a single scale FCN but it performed worse, likely due to smaller surrounding context for each pixel.

\subsection{Quadrilaterals}
\label{sec:quads}

Thresholding output of the FCN yields a binary image (Figure~\ref{fig:post_processing}b), which is converted to the coordinates of a quadrilateral around the main page region in the image.
While the pixel representation may be already useful for some applications (e.g., masking text detection regions), it can lack global and local spatial consistency due to the FCN predicting pixels independently based on local context.
Representing the detected page region as a quadrilateral fixes some errors in the FCN output and makes it easier for potential downstream processing to incorporate this information.

Figure~\ref{fig:post_processing} demonstrates the following post-processing steps that converts the binary per-pixel output of the FCN (after thresholding at 0.5 probability) to a quadrilateral region:
\begin{enumerate}
\item Remove all but the largest foreground components.
\item Fill in any holes in the remaining foreground component.
\item Find the minimum area oriented bounding rectangle using Rotating Calipers method~\cite{toussaint1983}.
\item Iteratively perturb corners in greedy fashion to maximize IoU between the quadrilateral and the predicted pixels.
\end{enumerate}
\label{sec:arch}

Step 1 helps remove any extraneous components (false positives) that were predicted as page regions.
Some of these errors occur because the FCN makes local classification decisions. 
Similarly, Step 2 removes false negatives.
In Step 3, we find a (rotated) bounding box for the main page region (OpenCV implementation~\cite{opencv}), but the bounding box encloses all predicted foreground pixels and is therefore sensitive to any false positive pixels that are outside the true page boundary.
This bounding box is used to initialize the corners for iterative refinement in Step 4.
At each refinement iteration, we measure the IoU of 16 perturbations of the corners (4 corners moved 1 pixel in 4 directions) and greedily update with the perturbation that has the highest IoU w.r.t.~the FCN output.
We stop the process when no perturbation improves IoU.
Post-processing is done on 256x256 images, so there may be quantization artifacts after the quadrilaterals are upsampled to the original size.

\subsection{\systemname{} Implementation Details}

\begin{table*}
\centering
\begin{tabular}{| c | c | c c c c c | c |}
  \hline  
   Dataset   & \# Images & \systemname{} & \systemname{}    & Full  & Mean   & GrabCut & Human     \\
             &           & (pixels)      & (quads)          & Image & Quad   &         & Agreement \\
  \hline
  CBAD-Train & 1635      & 0.968         & \textbf{0.971}   & 0.823 & 0.883  & 0.900   &  -        \\
  CBAD-Val   & 200       & 0.966         & \textbf{0.968}   & 0.831 & 0.891  & 0.906   &  0.978    \\
  CBAD-Test  & 200       & 0.972         & \textbf{0.974}   & 0.839 & 0.894  & 0.916   &  0.983    \\
  \hline  
  PMJT       & 140       & 0.936         & \textbf{0.943}   & 0.809 & 0.897  & 0.904   & 0.982     \\
  Saint Gall & 60        & 0.975         & \textbf{0.987}   & 0.722 & 0.809  & 0.919   & 0.993     \\
  Parzival   & 47        & 0.956         & \textbf{0.962}   & 0.848 & 0.920  & 0.925   & 0.989     \\
  \hline
\end{tabular}
\caption{mIoU of \systemname{} and baseline systems.  All rows used the same \systemname{} model trained on CBAD-train.  \systemname{} (pixels) is the pixel segmentation of our proposed method taken after step 2 of post-processing (see Section~\ref{sec:quads} for details).}
\label{tab:main_results}
\squeezeup
\end{table*}

We implemented the FCN part of \systemname{} using the popular deep learning library Caffe~\cite{jia2014}.
The dataset used for training is detailed in Section~\ref{sec:dataset}.
For preprocessing, color images are first resized to 256x256 pixels and pixel intensities are shifted and scaled to the range $[-0.5,0.5]$.
For ground truth, we label each pixel inside the image's annotated quadrilateral as foreground and all other pixels as background.
The ground truth images are also resized to 256x256.

While all input images yield a probability map of the same size, we used 256x256 images in training and evaluation.
While a larger input sizes could lead to slightly higher segmentation accuracy, we achieve good results with the computationally faster 256x256 size.
Initial experiments with 128x128 inputs were less accurate.

To train the FCN, we used Stochastic Gradient Descent for 15000 weight updates with a mini-batch size of 2 images.
We used an initial learning rate of 0.001, which was reduced to 0.0001 after 10000 weight updates.
We used a momentum of 0.9, L2 regularization of 0.0005, and clipped gradients to have an L2 norm of 10.
We trained 10 networks and used the validation set to select the best network for the results reported in Section~\ref{sec:results}.

\section{Dataset}
\label{sec:dataset}

Our main dataset is the ICDAR 2017 Competition on Baseline Detection (CBAD) dataset~\cite{cbad}, which are handwritten documents with varying levels of layout complexity (see Figure \ref{fig:example_problems}a,b,c).
We combined both tracks of the competition data and separated the images into training, validation and test sets with 1635, 200, and 200 images respectively.
We train \systemname{} on the training split of CBAD and evaluate on the validation and test splits of the same dataset.

We also evaluated on a subset of the CODH PMJT dataset\footnote{\url{http://codh.rois.ac.jp/char-shape/}}, and on all images from the Saint Gall and Parzival datasets. 
The PMJT (Pre-Modern Japanese Text) dataset is taken from the Center for Open Data in the Humanities (CODH)~\cite{pmjt} and consists of handwritten literature (see Figure \ref{fig:example_problems}d) and some graphics.
We randomly sampled 10 pages from each the collections, excluding ID 20003967, to create an evaluation set of 140 images.
The Saint Gall dataset~\cite{Fischer2011} is a collection of 9th century manuscripts put together by the FKI: Research Group on Computer Vision and Artificial Intelligence. 
The Parzival dataset~\cite{Fischer2012} is also put together by FKI and consists of 13th century Medieval German texts.  

We also trained and evaluated \systemname{} on a private collection of Ohio death records\footnote{Data provided by FamilySearch}, having a training set of 800 images, and validation and testing sets of 100 images each. 
This dataset, while relatively uniform in the types of documents, presents a unique challenge of overlay documents (see Figure \ref{fig:example_problems}e).  
While much of the document underneath an overlay is visible, much is still occluded, thus it is most desirable to localize just the overlay, which is a task that has not received much attention in the literature.
These overlays represent approximately 12\% of the images in the dataset.

\subsection{Ground Truth Annotation}

Our ground truth annotations consist of quadrilaterals encompassing the pages fully present in an image, not including any partial pages or page edges (if possible).
We chose to use quadrilaterals rather than pixel level annotations as most pages are quadrilateral-shaped, and it is much faster to annotate polygon regions than pixel regions.
Regions were manually annotated using an interface where the annotator clicks on each of the four corner vertices.
In the case of multiple full pages, the quadrilateral encloses all pages present regardless of their orientation to each other.
This typically occurs when two pages of a book are captured in a single image.

In order to give an upper bound on expected automated performance, we measured human agreement on the quadrilateral regions of our datasets.
A second annotator provided region annotations for validation and test sets.
To measure human agreement, this second set of regions was treated the same as the output of an automated system and scored w.r.t. the first set of annotations.

\section{Results}
\label{sec:results}

\begin{figure*}
\subfloat[]{\includegraphics[width=0.24\textwidth,height=0.22\textwidth]{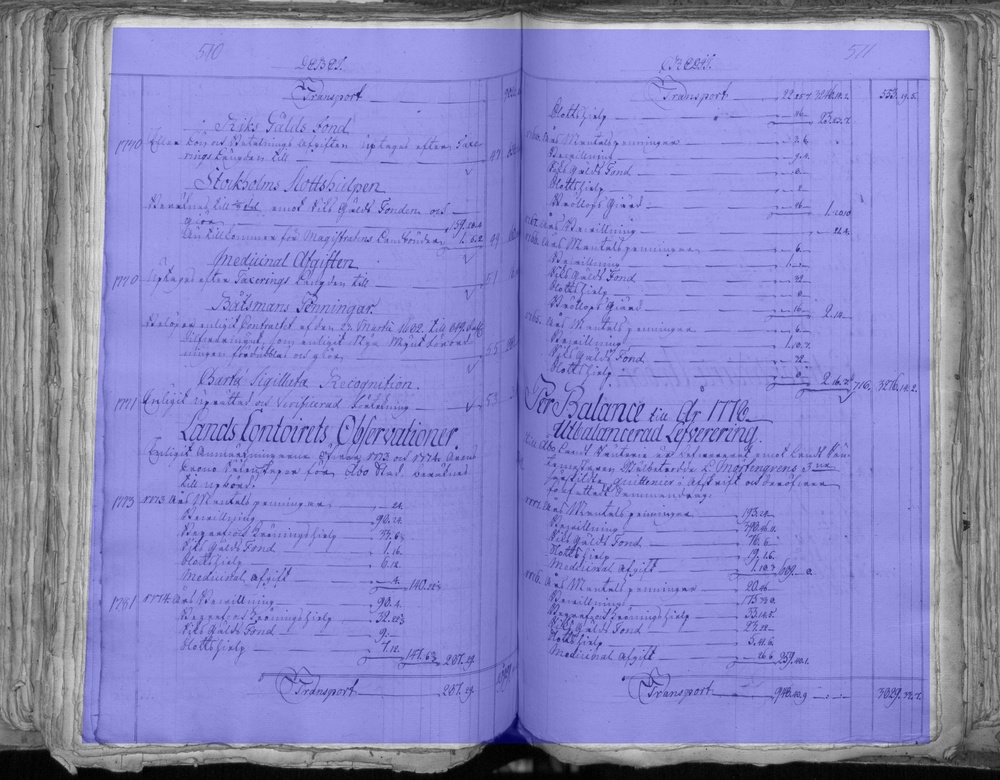}} \hspace{3pt}
\subfloat[]{\includegraphics[width=0.175\textwidth,height=0.22\textwidth]{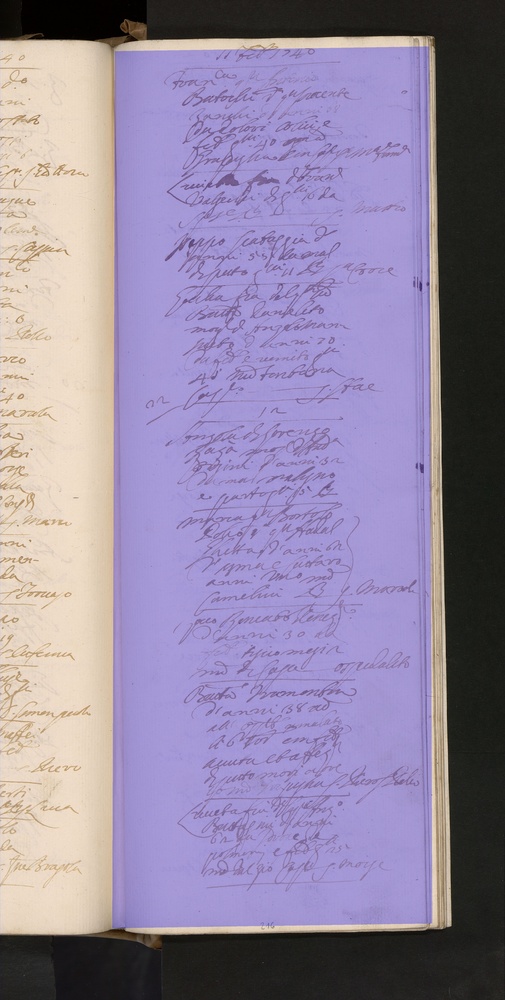}} \hspace{3pt}
\subfloat[]{\includegraphics[width=0.175\textwidth,height=0.22\textwidth]{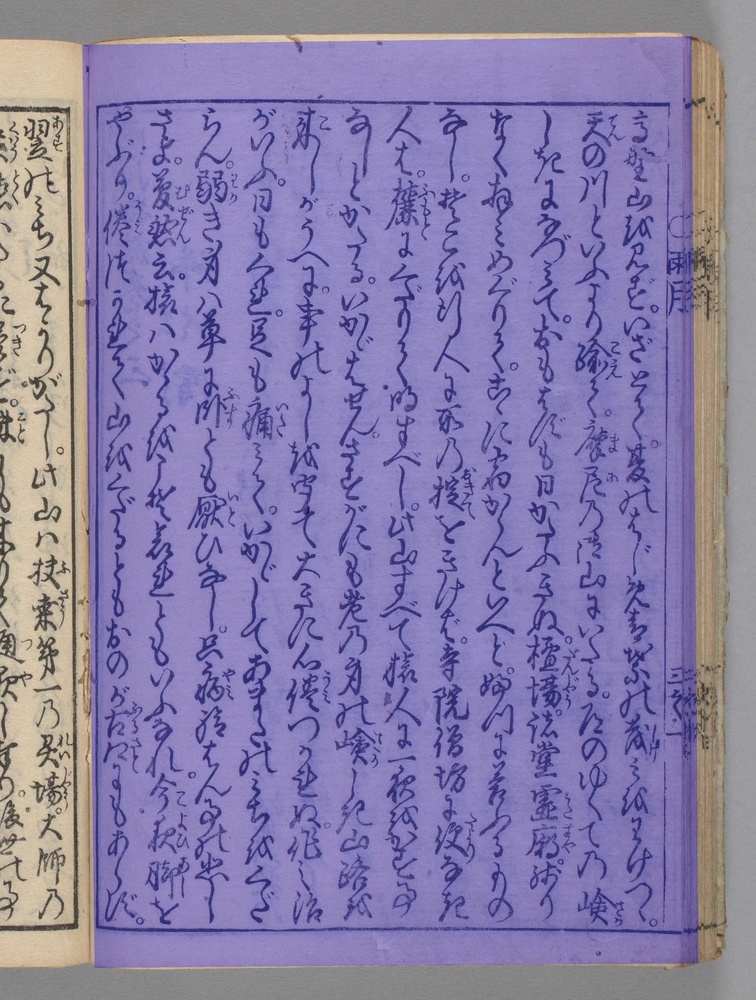}} \hspace{3pt}
\subfloat[]{\includegraphics[width=0.175\textwidth,height=0.22\textwidth]{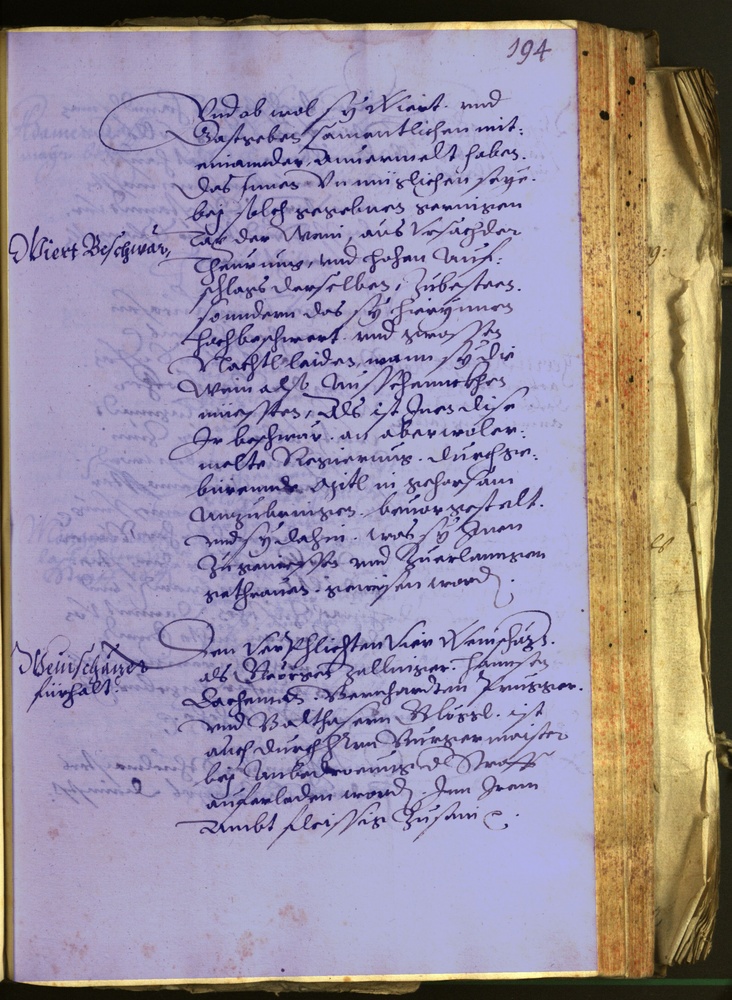}} \hspace{3pt}
\subfloat[]{\includegraphics[width=0.175\textwidth,height=0.22\textwidth]{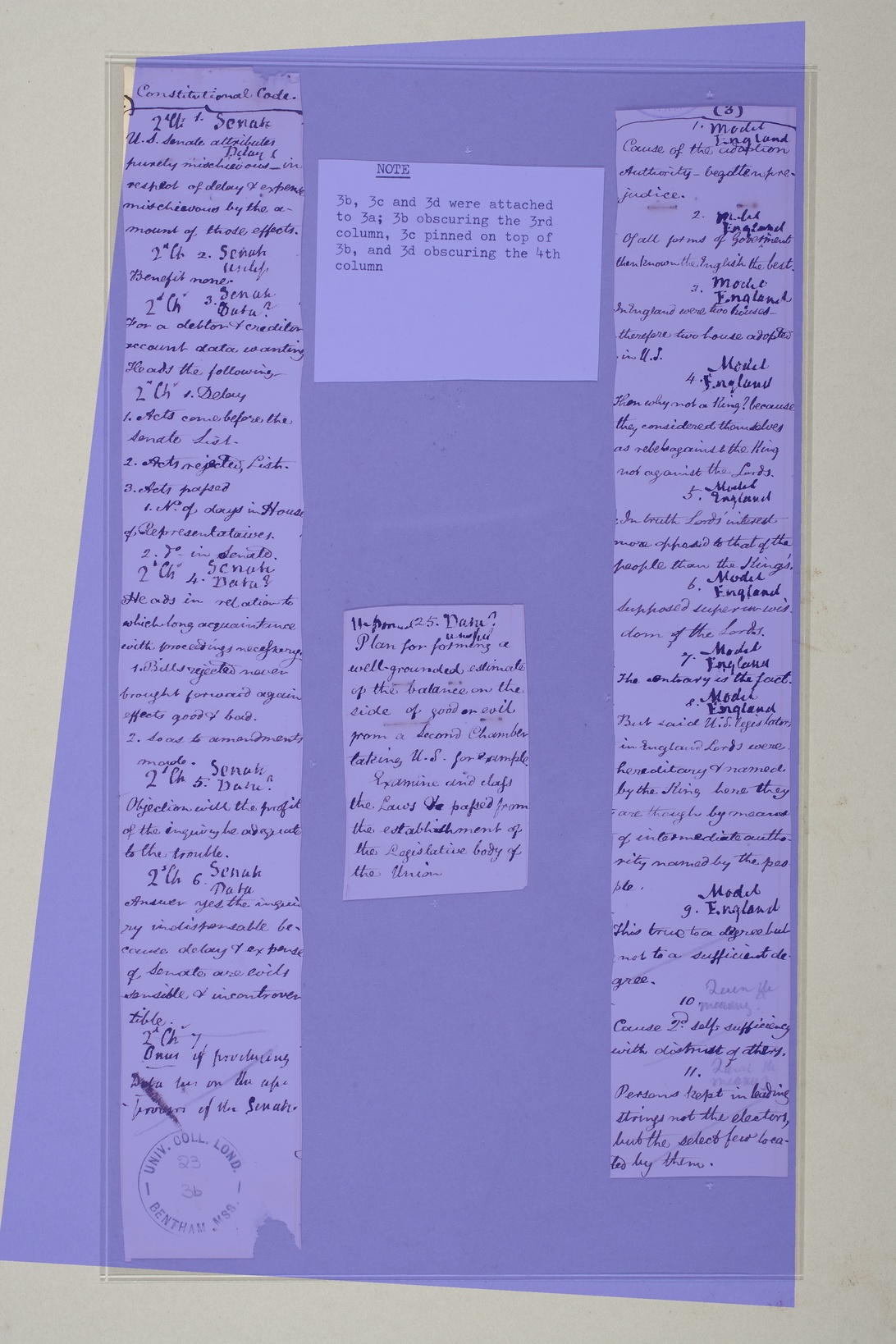}}

\caption{Example segmentations produced by \systemname .  (e) is an example failure case.}
\label{fig:results}
\end{figure*}

In this section, we quantitatively and qualitatively compare \systemname{} with baseline systems and with human annotators.
To evaluate system performance, we use Intersection over Union (IoU) averaged over all images in a dataset (mean IoU).
We chose mIoU as our metric because it is commonly used to evaluate segmentation task.

\subsection{Baselines systems}

We compare \systemname{} with three baseline systems: full image, mean quadrilateral, and GrabCut~\cite{rother2004}.
For the full image baseline, the entire image is predicted as the main page region.

The mean quadrilateral can be computed as 
\begin{equation}
\bar{x}_n = \frac{1}{N} \sum_{i=1}^N \frac{x_{in}}{w_i} \;\;\; \bar{y}_n = \frac{1}{N} \sum_{i=1}^N \frac{y_{in}}{h_i}
\end{equation}
where the mean quadrilateral is $(\bar{x}_1, \bar{y}_1, \dots, \bar{x}_4, \bar{y}_4)$, the $i$th annotated quadrilateral is $(x_{i1}, y_{i1}, \dots, x_{i4}, y_{i4})$, $N$ is the number of training images, $w_i$ and $h_i$ are respectively the width and height of the $i$th image.
The predicted quadrilateral for this baseline only relies on the height and width on the test image.
For the $j$th image, the prediction is $(w_j \bar{x}_1,$ $h_j \bar{y}_1, \dots, w_j \bar{x}_4, h_j \bar{y}_4)$.
Our mean quadrilateral was computed from the CBAD training split. 

For the GrabCut baseline, we used the implementation in the OpenCV library~\cite{opencv}.
For an initial object bounding box, we include the whole image except for a 5 pixel wide border around the image edges.
Unlike other baselines and the full \systemname{} results, GrabCut outputs a pixel mask (i.e., not a quadrilateral).
Therefore, it is directly comparable to \systemname{} before we extract a quadrilateral.

\subsection{Overall Results}

We trained \systemname{} on CBAD-train and tested it and the baseline methods on 4 datasets of historical handwritten documents.
Numerical results are shown in Table~\ref{tab:main_results}, and some example images are shown in Figure~\ref{fig:results}.
For all datasets, except CBAD-train, we manually annotated each image twice in order to estimate human agreement on this task.

On all datasets, the full \systemname{} system performed the best of all automated systems and strongly outperformed the baseline methods.
Notably, outputting quadrilaterals improves the pixel segmentation produced by the FCN, which shows that outputting a simpler region description does not decrease segmentation quality.
There is little difference in the results for the different splits of CBAD, which indicates that \systemname{} does not overfit the training images.
Performance is highest on Saint Gall because the border noise is largely limited to a black background.
On PMJT, \systemname{} performed worst and most errors can be attributed to incorrectly identifying page boundaries between pages, perhaps because the Japanese text is vertically aligned.

The full image baseline performs worst as it simply measures the average normalized area of the main page region.
With the exception of Saint Gall, GrabCut only marginally outperforms the mean quadrilateral.
As GrabCut is based on colors and edges, it often fails to exclude partial page regions (e.g., Figure~\ref{fig:example_seg}) and sometimes labels dark text regions the same as the dark background.
A few images in CBAD are well cropped and contain only the main page region, which is problematic for GrabCut because it will always attempt to find two distinct regions.
In contrast, once trained, \systemname{} can classify an image as entirely the main page region if it does not contain border noise.

\subsection{Comparison to Human Agreement}
The last column of Table~\ref{tab:main_results} shows the performance of a second annotator scored w.r.t. the original annotations.
This performance captures the degree of error, or ambiguity, inherent with human annotations on this task, a level of performance which would be difficult for any automated system to surpass.
For CBAD, \systemname{} is roughly 1\% below human agreement, which indicates the network's proficiency at the task.
On the PMJT dataset there is a larger gap between automated and human performance and indicates that there is still room for improvement.

The human agreement results in Table~\ref{tab:main_results} show the agreement between two annotators.
We also measured the agreement of the same annotator labeling the same images on a different day.
The same-annotator mIOU are 99.0\% and 98.4\% for CBAD-test and CBAD-val respectively.
These are slightly higher than the mIOU of 98.3\% and 97.8\% obtained by a different annotator.
This highlights the inherit ambiguity in labeling the corners of the main page region.

\subsection{Overlay Performance}

\begin{figure}
\subfloat[Overlayed]{\includegraphics[width=0.22\textwidth]{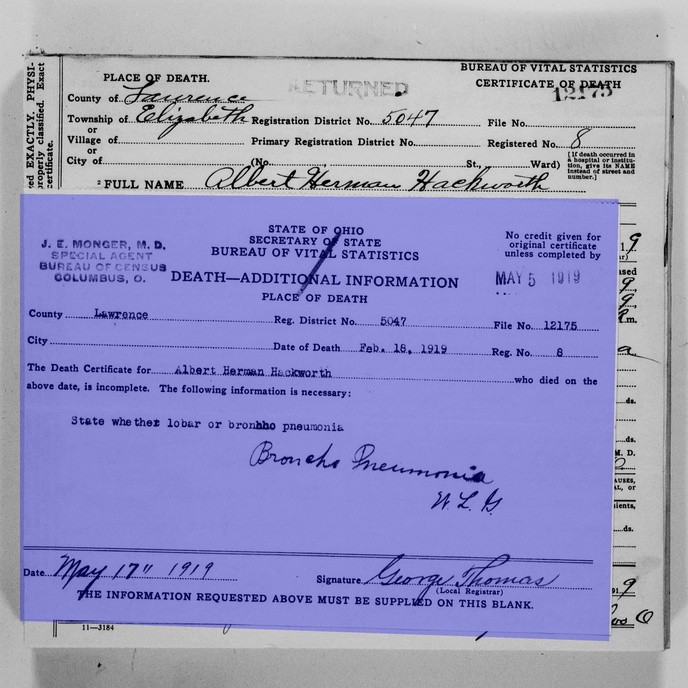}} \hspace{5pt}
\subfloat[Document Beneath]{\includegraphics[width=0.22\textwidth]{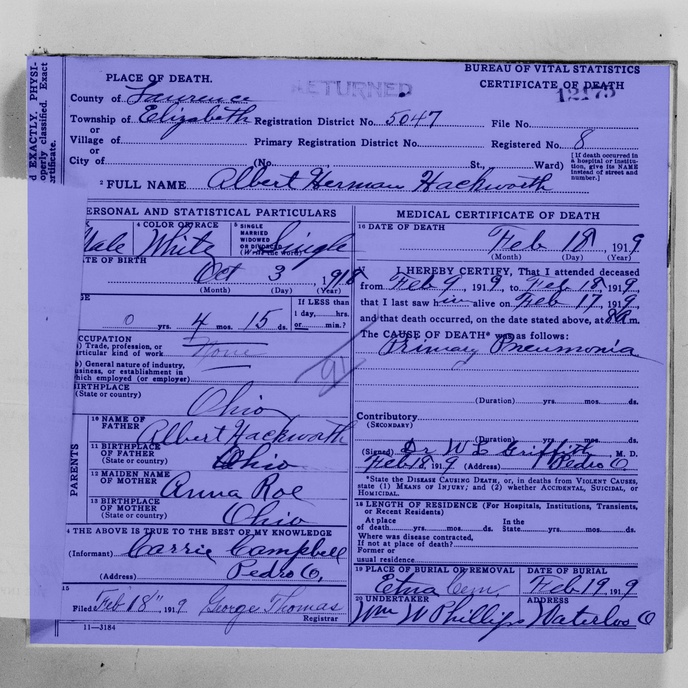}}

\caption{\systemname{} predictions for overlayed document.  The document underneath the overlay in (a) is the same document without the overlay in (b)}
\label{fig:overlay}
\end{figure}

\begin{table}
\centering

\begin{tabular}{| c | c | c c |}
  \hline
   Dataset   & \# Images & FCN      & FCN     \\
             &           & (pixels) & (quads) \\
  \hline
  Ohio-Train & 800       &  0.973   &  0.979  \\
  Ohio-Val   & 100       &  0.970   &  0.977  \\
  Ohio-Test  & 100       &  0.967   &  0.976  \\
  \hline
\end{tabular}
\caption{mIoU results of \systemname{} trained on Ohio death certificates.}
\label{tab:ohio_results}
\squeezeup
\end{table}

We also trained \systemname{} on a private dataset of Ohio death records.
This dataset has several images where one document is overlayed on top of another document (e.g., Figure~\ref{fig:example_problems}e), which creates particularly challenging textual noise.
Table~\ref{tab:ohio_results} shows results on this dataset.

In Figure~\ref{fig:overlay}, we show predicted segmentation masks for two images, which together show that \systemname{} correctly segments the overlayed image when present.
Figure~\ref{fig:overlay}a contains a document overlayed on top of the document shown in Figure~\ref{fig:overlay}b.
With the overlayed document, \systemname{} segments only the overlayed document, but when the overlayed document is removed, it segments the document underneath from the background.

\section{Conclusion}

We have presented a deep learning system, \systemname{}, which removes border noise by segmenting the main page region from the rest of the image.
An FCN first predicts a class for each input pixel, and then a quadrilateral region is extracted from the output of the FCN.
We demonstrated near human performance on images similar to the training set and showed good performance on images from other collections.
On an additional collection, we showed that \systemname{} can correctly segment overlayed documents.

\bibliographystyle{ACM-Reference-Format}
\bibliography{bib}

\end{document}